\documentclass[journal]{IEEEtran}
\usepackage{multirow}

\ifCLASSINFOpdf
  \usepackage{amsmath,graphicx}
  \graphicspath{{./figs/}}
  \DeclareGraphicsExtensions{.pdf,.jpg,.jpeg,.png}
\else
\fi
\usepackage{array}

\ifCLASSOPTIONcompsoc
 \usepackage[caption=false,font=normalsize,labelfont=sf,textfont=sf]{subfig}
\else
 \usepackage[caption=false,font=footnotesize]{subfig}
\fi

\begin{document}
%
\title{Hallucination of speech recognition errors\\ with sequence to sequence learning}
%
%
%

\author{Prashant~Serai,~\IEEEmembership{Student Member,~IEEE,}
        Vishal~Sunder,~
        and~Eric~Fosler-Lussier,~\IEEEmembership{Senior Member,~IEEE}
\thanks{Manuscript received Xyzember XX, YYYY; revised Xyzember XX, YYYY; accepted Xyzember XX, YYYY. This research was supported in part by the National Science Foundation under grant ABC-XXXXXXX and in part by the Ohio Supercomputer Center. The associate editor coordinating the review of this manuscript and approving it for publication was Dr. Abc Def. \emph{(Corresponding author: Prashant Serai.)}}
\thanks{The authors are with the Department
of Computer Science and Engineering, The Ohio State University, Columbus,
OH 43210 USA e-mail: serai.1@osu.edu, sunder.9@osu.edu, fosler@cse.ohio-state.edu.}
}

\markboth{IEEE Transactions on Audio, Speech, and Language Processing~Vol.~xx, Xyzember~yyyy}%
{Serai \MakeLowercase{\textit{et al.}}: Hallucination of speech recognition errors with sequence to sequence learning}

\IEEEpubid{0000--0000/00\$00.00~\copyright~2021 IEEE}

\maketitle

\begin{abstract}

Automatic Speech Recognition (ASR) is an imperfect process that results in certain mismatches in ASR output text when compared to plain written text or transcriptions. When plain text data is to be used to train systems for spoken language understanding or ASR, a proven strategy to reduce said mismatch and prevent degradations, is to hallucinate what the ASR outputs would be given a gold transcription. Prior work in this domain has focused on modeling errors at the phonetic level, while using a lexicon to convert the phones to words, usually accompanied by an FST Language model. We present novel end-to-end models to directly predict hallucinated ASR word sequence outputs,  conditioning on an input word sequence as well as a corresponding phoneme sequence. This improves prior published results for recall of errors from an in-domain ASR system's transcription of unseen data, as well as an out-of-domain ASR system's transcriptions of audio from an unrelated task, while additionally exploring an in-between scenario when limited characterization data from the test ASR system is obtainable. To verify the extrinsic validity of the method, we also use our hallucinated ASR errors to augment training for a spoken question classifier, finding that they enable robustness to real ASR errors in a downstream task, when scarce or even zero task-specific audio was available at train-time.




\end{abstract}


\IEEEpeerreviewmaketitle

\section{Introduction}

\IEEEPARstart{F}{or} several decades the {\em speech-text data imbalance} has been a significant factor in the impedance mismatch between spoken language processing systems and text-based language processing systems. Use of speech in artificial intelligence applications is increasing, however there is not always enough semantically labelled speech for individual applications to be able to build directly supervised spoken language understanding systems for them. On the other hand, wide domain cloud based automatic speech recognizer (ASR) systems are trained on a lot of data, and even as black boxes to the developer, they are able to transcribe speech to text with a lower error rate (under certain circumstances). As the availability of text resources for training the natural language understanding (NLU) system for many tasks far exceed the amount of available transcribed speech, many end developers of spoken language understanding systems utilize ASR systems as an off-the-shelf or cloud-based solution for transcribing speech and cascade them with NLU systems trained on text data~\cite{tsvetkov2014augmenting,stiff2019improving,namazifar2020warped}.

The text obtained from ASR typically contains errors whether resulting from artifacts or biases of the speech recognizer model, its training data, etc., or from inherent phonetic confusibilities that exist in the language being recognized (e.g., homophonic or near homophonic sets of words). When off-the shelf ASR systems are deployed in technical domains such as medical use-cases, the domain mismatch can increase the word error rate (WER) of state-of-the-art systems to as much as ~40\% in some cases~\cite{mani2020asr}; even when word error rates are lower, the semantic changes introduced by the errors can critically affect the meaning of the transcripts for downstream tasks in a manner that is much severe than typed text modalities~\cite{zhou2018analysis}.

\IEEEpubidadjcol

In order to alleviate the adverse impact of ASR errors on NLU systems, one approach is to ``speechify'' the original input text for training an NLU system while treating it as intended spoken text. In this strategy, the NLU system is made to observe an input that contains the kind of errors expected from ASR at test time, and thus can learn to be robust to them. A crucial question is: can we predict the output behavior of an ASR system from intended spoken text, including when the system is a black box for the developer? Prior work, described in section \ref{sec:prior_work}, has looked at approaching the task of error prediction by building models of phoneme confusability. Approaches in this category generally rely upon an FST decoding graph comprised of Pronunciation and Language Models to translate hallucinated errors from phonemes to words, thus the prediction is not optimized end to end. Additionally, prior work has been limited in the exploitation of context (at the phoneme or word levels) into the prediction of errors made by the ASR systems.

Our previous work explored sequence to sequence learning to model phoneme confusability in a context-dependent manner, which resulted in improved recall of ASR errors when combined with a confusion matrix sampling technique~\cite{serai2019improving}, however we still relied upon an FST decoding graph to translate errors to a word sequence space. In this work, we hypothesize that the folding of the pronunciation and language modeling ability of the decoding graph, along with confusability modeling into a single network can enable these modeling abilities to be jointly optimized for error prediction, and allow better interplay between the models. Our novel approach uses sequence to sequence learning to directly predict hypothesized ASR outputs from intended spoken text.

A key ingredient in building a machine learning model to predict the errors made by an ASR system is: data about the kinds of ASR errors made by the system. In this respect, the use of cloud-based ASR systems also brings an additional challenge i.e., the lack of publicly available error-characterization data. In prior work, we treated the task of predicting errors made by cloud based systems only as an out-of-domain task. However, we reason that limited characterization data may be collected from time to time, and thus this out-of-domain task need not be completely out-of-domain too. In this paper, we investigate the effect of passing some speech from a standard corpus through a cloud based ASR system to finetune an error prediction model for such a black box recognizer.

This study extends preliminary results presented in~\cite{serai2020end}, where we explored models that directly translated word sequences of intended spoken text to word sequences of hypothesized ASR output. While these word-level end to end models allowed for an improved overall recall of ASR errors, we found that they would not recall some errors that a phonetic confusion matrix model was able to recall, suggesting complementary information in the word and phonetic representations.


In this paper, along with the aforementioned word-level model, we present a dual encoder model for error prediction that can look at both word and phoneme sequence representations of input text to further improve the fidelity of hallucinated errors. We also expand on our preliminary experiments and evaluation in several ways. For evaluation on in-domain ASR, we look at a larger test set in addition to evaluating on a smaller one for comparability to prior work.  For out-of-domain ASR such as cloud-based systems, along with evaluating on read speech versions of chatted dialog turns, in this paper we include results on a dataset of realistic spoken dialog turns, looking at multiple word error rate settings, for an intrinsic as well as extrinsic evaluation. Finally, we present additional experiments in a practical middle-case where domain-specific ASR training data is available but only to a limited amount.



\section{Prior Work}
\label{sec:prior_work}


Traditionally, approaches to the task of predicting or hallucinating speech recognition errors have characterized word errors as indirectly resulting out of phonetic substitutions, insertions, or deletions. A general framework in this direction was described by Fosler-Lussier et al.~\cite{fosler2005framework} wherein they built a matrix of how often each phoneme in input text was confused by the recognizer for each possible sequence of zero or more phonemes, and cast it as a Weighted Finite State Transducer (WFST) graph. Amongst ideas for developing a confusion model from the internals of an ASR system when accessible, Anguita et al.~\cite{anguita2005detection} looked at directly determining phone distances by looking inside the HMM-GMM acoustic model of a speech recognizer. Jyothi and Fosler-Lussier~\cite{jyothi2009comparison} combined the two aforementioned ideas and extended it to predict complete utterances of speech recognized text. Tan et al.~\cite{tan2010automatic} explored the idea that the confusion characteristics of a phoneme can be vary based on other phonemes in its context, and used a phrasal MT model to simulate ASR, but only evaluating the 1-best word sequence of the final output. Sagae et al.~\cite{sagae2012hallucinated} and Shivakumar et al.~\cite{shivakumar2019learning} considered word level phrasal MT modeling for error prediction but did not combine it with phonetic information, or directly evaluate the fidelity of predicted errors. Our prior work~\cite{serai2019improving} took the framework of Fosler-Lussier et al. with it's applicability to black box systems, and investigated the benefit of introducing contextual phonetic information through a neural sequence to sequence model, along with introducing a sampling based paradigm to better match the stochasticity of errors and confidence of neural network acoustic models.

ASR error modeling has also been used to train language models discriminatively such that they complement the shortcomings, i.e., error characteristics of ASR models and help prevent errors where possible. Jyothi and Fosler-Lussier~\cite{jyothi2010discriminative} applied their aforementioned error prediction model trained from ASR behavior on a certain dataset to improve WER on the same dataset. Kurata et al.~\cite{kurata2011training} applied an error prediction model trained from ASR characteristics on one dataset to improve WER on another dataset. Sagae et al.~\cite{sagae2012hallucinated} tried different methods for error prediction for discriminative training of a language model, and found that modeling confusability amongst phoneme phrase cohorts i.e., sequences of phonemes instead of individual phonemes helped obtain a larger improvement in WER, showing a benefit in modeling errors in a contextual manner. Shivakumar et al.~\cite{shivakumar2019learning} explored modeling confusability  at the level of phrases of words, and improved WER in a ASR system with a hybrid DNN-HMM acoustic model.

Knowledge of ASR errors has been used in training of NLU for various spoken language understanding tasks.  Tsvetkov et al.~\cite{tsvetkov2014augmenting} improve a phrasal machine translation system's response to spoken input by augmenting phrases in it's internal tables with variants containing hallucinated ASR errors derived from a phonetic confusion matrix approach. Ruiz et al.~\cite{ruiz2015adapting} construct a spoken machine translation system that conditions on phoneme sequence inputs which are generated with hallucinated ASR errors at train time to build robustness to their nature. Stiff et al.~\cite{stiff2019improving} utilized our aforementioned sampling based phonetic confusion matrix approach and randomly chose to hallucinate ASR on typed text input to an NLU system at train time to improve its performance on a test set with real ASR errors. Rao et al.~\cite{rao2020speech} improved their NLU classifier's robustness to ASR errors by conditioning it on ASR hidden states instead of direct text to expose it to ASR confusability information, focusing on a scenario where all training data for NLU was in the spoken domain.

\section{System Description}

We use convolutional sequence to sequence models~\cite{gehring2017convolutional} for the purpose of translating true text (gold transcripts free from ASR errors) to recognized text (transcription hypotheses with hallucinated ASR errors).

\subsection{Word level ASR prediction}
\label{ssec:word2word}
\begin{figure}[t]
    \centering
    \includegraphics[width=0.5\textwidth]{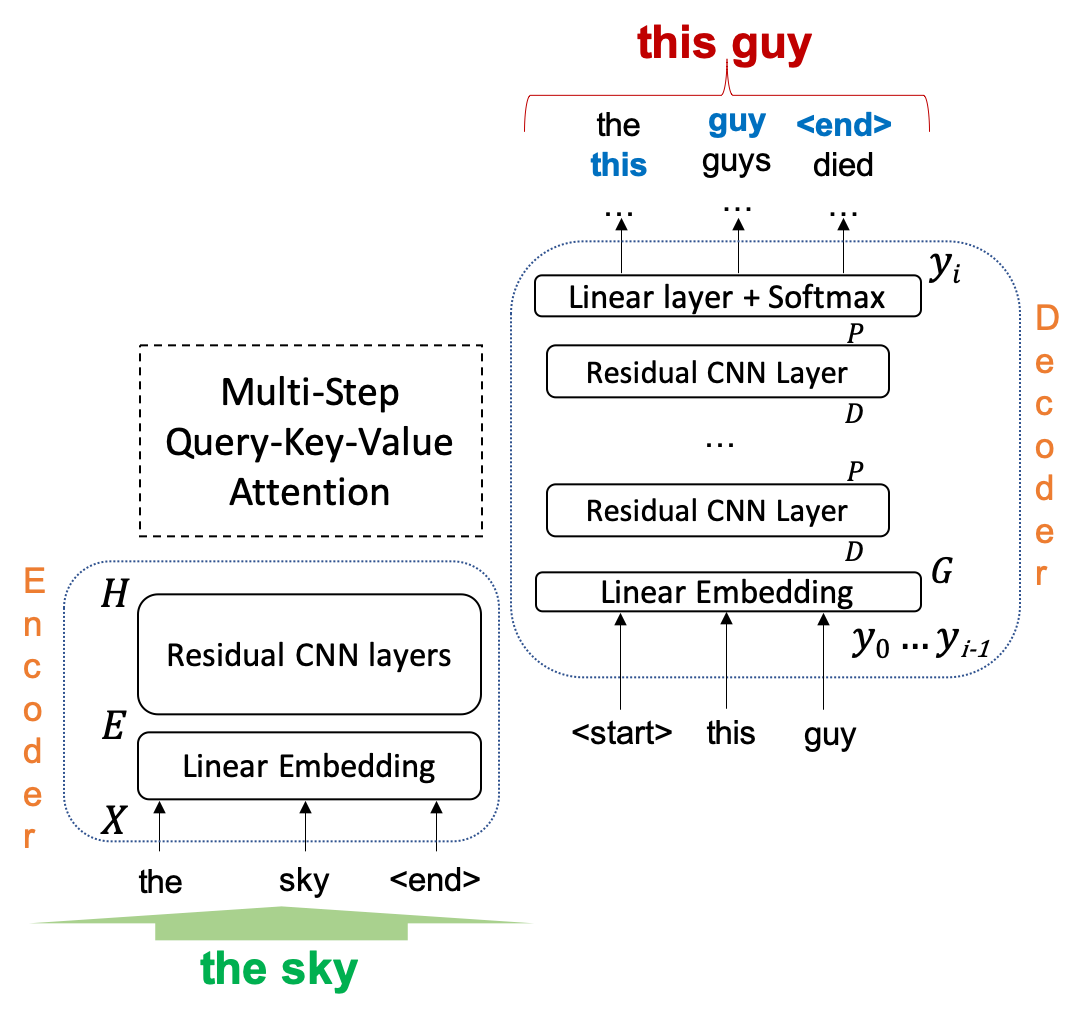}
    \caption{Architecture for the word level ASR error prediction model: The encoder is fed the text transcript as a padded word sequence. The decoder outputs one word at a time, conditioning upon the sequence of previously outputted words. This is used to construct an output word sequence through one of our decoding schemes.}
    \label{fig:word2word}
\end{figure}

The architecture for the word level ASR prediction model is shown in Figure \ref{fig:word2word}. An encoder takes a word sequence representation of the true text $X=x_1,\dots,x_n$ as input, and embeds it into a sequence of 256-dimensional vector representations (combined with position embeddings) $E=e_1,\ldots,e_n$. A stack of four residual CNN layers~\cite{he2016deep} transforms $E$ into a final hidden representation $H=h_1,\ldots,h_n$. Both the hidden representation $H$ and the embedded input $E$ are provided to an attention mechanism.

The decoder is comprised of three residual CNN layers. The decoder takes as input the sequence of predicted words prior to the current timestep, and embeds them into a sequence of vector representations $G=g_1,\ldots,g_{i-1}$, we use 256 dimensional embeddings here as well. Along with these embeddings, each decoder layer also conditions upon an attended representation from the encoder derived through the mechanism explained below. The output of the final layer is passed through a linear transformation followed by a softmax, to give a probability distribution over the target vocabulary at step $i$. Cumulatively, this model has 37M parameters.

\subsubsection{Attention Mechanism}
\label{sssec:word2word_attn}
For every decoder layer $l$ with input $D (d_{l,1},\ldots,d_{l,m})$ and output $P (p_{l,1},\ldots,p_{l,m})$, the attention computation can be expressed in a query-key-value formulation~\cite{vaswani2017attention}, wherein an output is calculated as a weighted sum of \textit{value vectors}, with the weights determined as a function of the \textit{query vector} and respective \textit{key vectors} corresponding to the value vectors.

For timestep $i \in \{1 \ldots m\}$ of the decoder, the query vector for layer $l$ is the combination of the current decoder state at timestep $i$ at the output of layer $l$, and embedding of the target predicted at the previous timestep, $g_{i-1}$.
\[q_{l,i} = W_l * p_{l,i} + b_l + g_{i-1}\]

From timestep $j\in \{1 \ldots n\}$ of the encoder, the value vector is computed from the encoder representation i.e.\ by a sum of the final hidden representation and the input embedding at that timepoint, whereas the key vector is just the final hidden representation.
\[k_j = h_j\]
\[v_j = h_j + e_j \]

The attention weight matrix from layer l is computed by a softmax over the product of the query and key vectors.
\[a_{l,i,j} = \dfrac{\exp(q_{l,i} * k_{j})} {\sum_{t} \exp(q_{l,t} * k_{j})}\]

These weights are then used to compute the attended input to decoder layer l+1, from the value vectors.
\[d_{l+1,i} = \sum_{t} a_{l,i,t}*v_{i} \]

By letting $v_j$ be a combination of the $h_j$ and $e_j$, we believe $h_j$ is enabled to effectively focus on learning confusion modes and/or likelihoods for the word in the sequence, and let the information about the word itself be contained in $e_j$.

\subsubsection{Decoding Mechanisms}
We use the output of the decoder to construct an N-best hypothesis list for recognized text, comparing two methods for list construction. In the first method ({\em Beam Search Decoding}), we use a left-to-right beam search as applied to sequence to sequence networks~\cite{sutskever2014sequence}, tracking $B=256$ running hypotheses sequences at a time. We 
select the 100-best complete hypotheses 
based on the cumulative length-normalized sequence probability. Our second method is based on the success of sampling in prior work for error prediction~\cite{serai2019improving}: we investigate a sampling based decoding technique, wherein at each timepoint $i$, we sample a word from the target vocabulary based on the output probability distribution of the decoder ({\em Sampled Decoding}). For every timestep $i$, the input contains embeddings of the target words chosen from timestep $1\ldots i$. We generate a minimum of 250, and generate until we have 100 unique sequences, or hit a maximum of 1000 word sequence samples. If we obtain more than 100 unique sequences, we select the most frequently occurring 100.

\subsection{Incorporating Phonetics into ASR prediction}
For words and word-pairs where the model is unable to capture enough examples of possible recognitions or mis-recognitions, if we can make additional information about how each word sounds like (through the phonemes), the model could learn to ``backoff'' to the phoneme representation as needed. Thus, to improve generalizability and aid learning, we look at incorporating a phonetic representation of the true text as an additional input. 

\begin{figure}[t]
    \centering
    \includegraphics[width=0.5\textwidth]{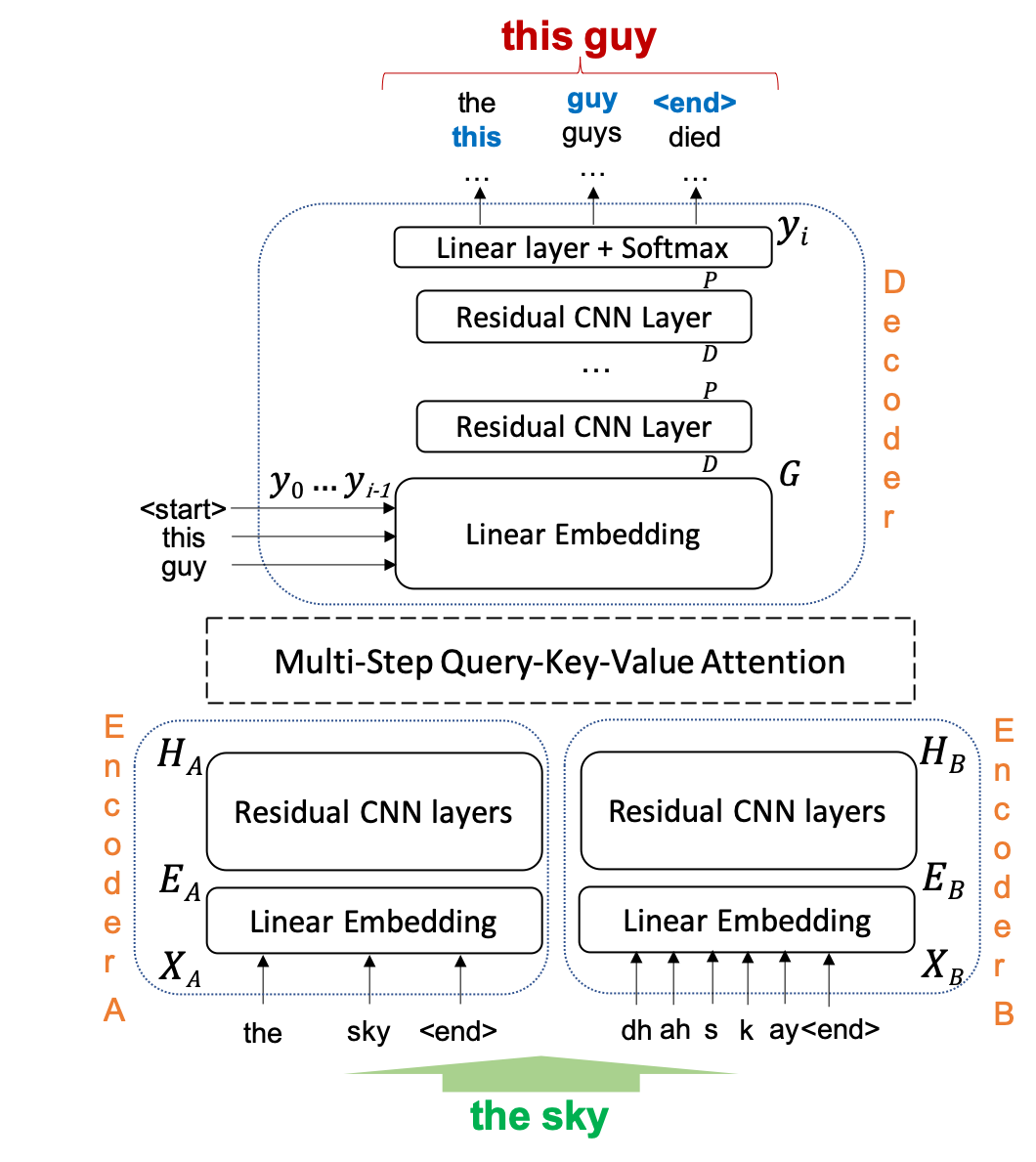}
    \caption{A dual encoder end-to-end model with a word sequence decoder conditioned on a word sequence encoder and a phoneme sequence encoder}
    \label{fig:dual_encoder}
\end{figure}

Accordingly, we propose a sequence to sequence model with two encoders and one decoder as shown in Figure \ref{fig:dual_encoder}. Encoder A takes a word sequence  corresponding to the true text whereas encoder B takes the phoneme sequence corresponding to the same. The decoder attends to both encoders to produce predictions for the recognized word sequence.

In this model, we use the same four layer decoder architecture as in the word level ASR prediction model, but in the encoders we use wider kernels and increase the number of layers, so as to account for phoneme sequences being longer than word sequences, while keeping the number of parameters comparable. Each encoder comprises of three residual convolutional layers with 64 filters and a kernel size of 11, followed by two residual layers with 128 filters and a kernel size of 7, and finally one residual layer with 256 filters and a kernel size of 5. Cumulatively, this model has ~38M parameters, which is comparable to the word level model. 

To allow the decoder to look at both word and phoneme sequence encoders, we propose a dual attention mechanism  detailed in \ref{ssec:dual_attn} below, and to further encourage it to learn to incorporate both sources of information, we introduce an encoder dropout mechanism as detailed in \ref{ssec:enc_dropout}. In limited experimentation, we also tried adding a second decoder with an auxiliary objective of predicting the phoneme sequence representation of the recognized text, but it did not seem to change the results much, as a result we did not explore it further.

\subsubsection{Dual Attention Mechanism}
\label{ssec:dual_attn}
We propose to adapt the attention mechanism from section \ref{sssec:word2word_attn} to two encoders. For every decoder layer $l$ with input $D (d_{l,1},\ldots,d_{l,m})$, output $P (p_{l,1},\ldots,p_{l,m})$, the attention computation can be expressed in a similar query-key-value formulation as follows.

For timestep $i \in \{1 \ldots m\}$ of the decoder, the query vector for layer $l$ corresponding to encoder $y$ is the combination of the current decoder state $p$ at timestep $i$ at the output of layer $l$, and embedding of the target predicted at the previous timestep, $g_{i-1}$.
\[q_{y,l,i} = W_l * p_{l,i} + b_l + g_{i-1}\]

From timestep $j\in \{1 \ldots n\}$ of encoder $y$, the value vector is computed from the corresponding encoder representation i.e.\ by a sum of the final hidden representation and the input embedding at that timepoint, whereas the key vector is just the final hidden representation.
\[k_{y,j} = h_{y,j}\]
\[v_{y,j} = h_{y,j} + e_{y,j} \]

The attention weight matrix from layer l is computed by a softmax over the product of the query and key vectors.
\[a_{y,l,i,j} = \dfrac{\exp(q_{y,l,i} * k_{y,j})} {\sum_{t} \exp(q_{y,l,t} * k_{y,j})}\]

These weights are then used to compute the attended input to decoder layer l+1, from the value vectors. The weighted representations from the heads attending to both the encoders are concatenated and then combined using a linear transformation.
\[v_{attended_{A,l,i}} = \sum_{t} a_{A,l,i,t}*v_{A,i}\]
\[v_{attended_{B,l,i}} = \sum_{t} a_{B,l,i,t}*v_{B,i}\]
\[d_{l+1,i} = W_{dual} * (v_{attended_{A,l,i}} \oplus v_{attended_{B,l,i}}) \]

\subsubsection{Encoder dropout}
\label{ssec:enc_dropout}

In our dual encoder model, we allow the decoder to attend to multiple encoders simultaneously, however, the decoder could learn to just use the information from one of the encoders and ignore the other. For example, in Figure \ref{fig:dual_encoder} the decoder, can learn to just focus on the words encoded by Encoder B and discard the phonetic information from Encoder A, thus defeating the dual attention mechanism. We propose an encoder dropout scheme to encourage the decoder to learn to focus on both encoders, by letting it have access to only one of the encoders at certain times.

For an encoder dropout factor $p_d \in \{0-1\}$, with probability $p_d$ we decide to drop exactly one of the two encoders picked at random. Specifically, for every example in a training batch:\\
1. With probability $p_d/2$, we drop encoder A in the following manner:
\[v_{attended_A} := 0 * v_{attended_A}\]
\[v_{attended_B} := 2 * v_{attended_B}\]
2. Else, with $p_d/2$ of the remaining probability, we drop encoder B in the following manner:
\[v_{attended_A} := 2 * v_{attended_A}\]
\[v_{attended_B} := 0 * v_{attended_B}\]
3. Else, with $1 - p_d$ of the remaining probability, we drop neither of $v_{attended_B}$ and $v_{attended_A}$, i.e., leave them both untouched.

For every example that one of the encoders is dropped, the other corresponding encoder's attended representation is multiplied by a factor of 2 to compensate for the additional input. Additionally, with the $1 - p_d$ chance of no dropout, we encourage the decoder to learn not only to attend to each encoder individually, but also learn to attend to both of them simultaneously. We apply this encoder dropout in addition to using conventional dropout at the output of every layer in the encoder and decoder.


\section{Data Preparation and Task Setups}
\begin{figure*}[tb]
    \centering
    \includegraphics[width=1.0\textwidth]{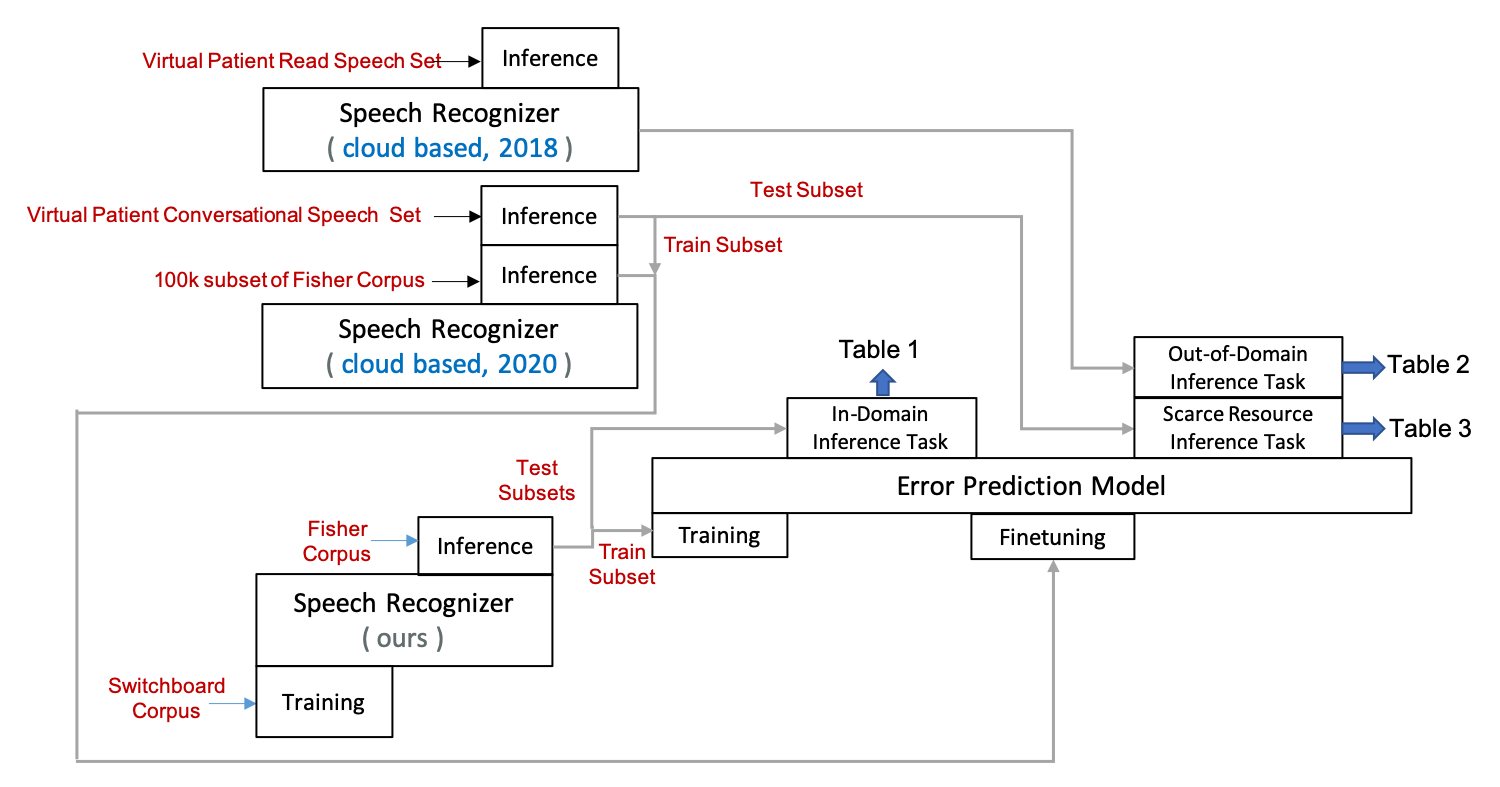}
    \caption{A data-flow schematic of how different sets are used for training, finetuning, and evaluation of the error prediction model.}
    \label{fig:train_finetune_flowchart}
\end{figure*}

The task of hallucination or prediction of errors is treated as a translation problem, from true text to recognized text. Figure \ref{fig:train_finetune_flowchart} shows a schematic of how various sets of data are used for training or evaluation of the error prediction systems, and the construction of those sets is described below. The primary training data is derived using the Fisher corpus, and an ``in-domain'' evaluation is performed on unseen examples from the same corpus and same ASR system observed at train-time. For an ``out-of-domain'' evaluation, we follow prior work to utilize a set based on data from The Ohio State University's Virtual Patient project (described in subsection \ref{ssec:vpdata}), where the ASR system and corpus are both unobserved at train-time. We  also conduct a ``scarce-resource'' evaluation with other data from the aforementioned Virtual Patient project, wherein we collect some examples of recognition with the test-time ASR system to make a ``finetuning set'' from the Fisher corpus as well as from the Virtual Patient project. Along with evaluating the quality of our hallucinated ASR hypotheses, we study the downstream impact of our hallucination; this ``extrinsic evaluation'' is performed on the Virtual Patient spoken question classification task.



\subsection{Fisher Data}
Fisher is a conversational telephonic speech corpus in English containing 2000 hours of audio data paired with human annotated transcriptions~\cite{cieri2004fisher}, segmented into 1.8 million odd utterances. We transcribe Fisher using multiple ASR systems, in order to create pairs of ``true text'' (human annotated or corrected) and ``recognized text'' (ASR transcribed), used for training and evaluation of the error prediction system.
\subsubsection{In-domain Set (Fisher\_base)}
\label{sssec:fisher_base}
Our primary source ASR system utilizes the Kaldi Switchboard recipe, training a DNN with the sMBR criterion, and decoding with a trigram language grammar trained exclusively on the text in the Switchboard corpus~\cite{vesely2013sequence}. We use this recognizer to obtain 1-best transcriptions for the 1.8 million odd utterances in the Fisher corpus at a roughly 30\% word error rate. The standard train split was used for training, and standard validation split for validation, for all versions of ASR hallucination models except for the ``only-finetune'' case in the scarce resource evaluation setting (\ref{table-vp2}). For testing the in the in-domain setting, the standard test split of 5000 examples was used in conjunction with a smaller randomly chosen subset of 500 examples used in prior work.

\subsubsection{Finetuning Set (Fisher\_finetune)}
\label{sssec:fisher_finetune}
Our secondary source is a commercially available cloud-based ASR system used in late 2020, that is the same as the one we intended to use for transcription in one version of our target spoken language understanding task; we do not have further access to the internals or the details of the training of this system. Since transcription requests to this system were rate-limited, and had a cost associated to them, we randomly selected a subset of 100k utterances from the training set of the Fisher corpus, corresponding to about 104 hours of audio. We used LDC's sph2pipe software to read and convert the audio corresponding to these selected utterances to wav files, and subsequently interpolated them to a sample rate of 16khz using Librosa~\cite{librosa} to match the input specification for the ASR. These resampled utterances were then transcribed using the ASR at a roughly 17\% word error rate. The resulting set was used for finetuning or training the ASR hallucination model in the post-finetune and only-finetune cases of the scarce resource evaluation setting, respectively (Table \ref{table-vp2}). It was also used in the finetuning of the error hallucination models used in the downstream evaluation setting (Table \ref{table-vp-downstream}). Except for the zero in-domain ASR case in the downstream evaluation setting, the finetuning set for the ASR hallucination model also included 4991 annotated and cloud-ASR transcript pairs from the ``training set'' portion of the Virtual Patient Conversational Speech Set (VP\_conv) described below, along with the set described in herein. 

\subsection{Virtual Patient Data}
\label{ssec:vpdata}
The virtual patient is a graphical avatar based spoken dialog system for doctors to practise interviewing patients (see figure \ref{fig:vp_gui}). The virtual patient is designed to be capable of answering a limited set of questions (that have fixed and pre-written answers), creating the task of question classification based on user input. The following are different sets of virtual patient data we use:

\begin{figure}[tb]
    \centering
    \includegraphics[width=0.45\textwidth]{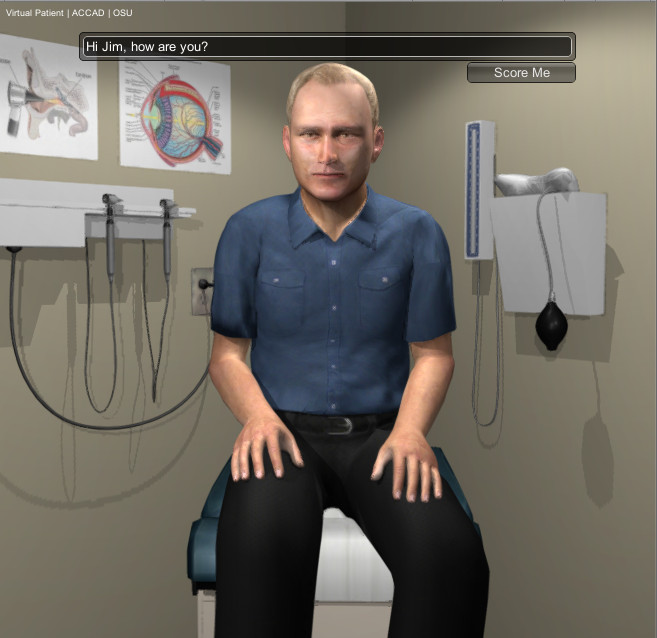}
    \caption{Graphical user interface for the Ohio State University’s Virtual Patient.}
    \label{fig:vp_gui}
\end{figure}

\subsubsection{Text Set (VP\_text)}
\label{sssec:vp_text}
The Virtual Patient Text Set consists of 259 type-written dialogues of users interacting with an older version of the Virtual Patient prior to incorporation of speech input~\cite{stiff2020self}. We use this data as part of the training set for the question classification model in the extrinsic evaluation. As the nature of this text is typed, there exists a mismatch with speech recognized text, and thus also a potential for ASR hallucination. The set contains a total of 6711 examples of user turns paired with human-annotated question/answer labels.

\subsubsection{Read Speech Subset (VP\_read)}
\label{sssec:vp_read}
To evaluate our error prediction model in an out-of-domain setting in a comparative manner to prior work~\cite{serai2019improving}, we utilize the read speech set. It consists of 756 utterances that were taken as a subset from the text set, read by volunteers, and transcribed with a commercially available cloud based ASR service in 2018, with a word error rate of slightly over 10\%~\cite{stiff2019improving}.

\subsubsection{Conversational Speech Set (VP\_conv)}
\label{sssec:vp_conv}
To evaluate our error prediction model in a realistic spoken dialog setting, we utilize data collected from a spoken dialog version of the Virtual Patient, where speech input from users was fed through a commercially available cloud based ASR service in late 2018, and the resulting natural language was passed to a question classifier that was a combination of a machine learning system trained on the text set \ref{sssec:vp_text} along with hand-crafted patterns. This contained 11,960 user turns over 260 conversations or dialogues. Human annotations were performed to obtain text transcripts (``true text'') as well as question/answer labels. This led to one set of pairs of ``true text'' and ``recognized text'', where the word error rate for these transcriptions from 2018 was calculated to be around 12.5\%.

The ASR transcriptions from the cloud based system used in 2018, with a word error rate of around 12.5\% formed one ``recognized text'' version of the data. However, it is important to understand how well the error prediction generalizes across ASR systems in cases where domain data is seen for finetuning; these should correspond to several points along the accuracy spectrum.  
%
%
We resampled the speech collected with the 2018 cloud-based system to 16KHz 
and
passed it through two more speech recognizers to create more versions of recognized text for this data. First, it was passed through a separate commercially available cloud-based ASR service in 2020 (identical to \ref{sssec:fisher_finetune}), this had a word error rate of 8.9\%. Second, it was passed through an ASR model trained on the Librispeech corpus using SpeechBrain~\cite{SB2021}. 
As there is a significant mismatch in terms of domain, style of speech, and vocabulary, the Librispeech-based system has a word error rate of 41.1\%, which serves as a ``worst case'' system.

For the purpose of our experiments, we randomly split the 260 dialogues into a training set of utterances from 100 dialogues (4991 turns), a validation set of 60 dialogues (1693 turns), and a test set consisting of the remaining 100 dialogues (5118 turns).

For training and validating the spoken question classification model, the human annotated transcripts of the inputs along with labels are used in the zero ASR data case, whereas in the case for some ASR data being available, the cloud-ASR transcripts of the inputs from 2020 are additionally employed. For testing the question classification model, we look at transcripts from all aforementioned ASR systems as well as human annotated transcripts.

For the ASR hallucination model, the cloud-ASR transcripts from 2020 are  used for training, validation, and testing in the post-finetune and only-finetune cases of the scarce-resource evaluation setting (Table \ref{table-vp2}). 

\subsection{Data Preprocessing}
\label{data_preprocessing}
The true text and recognized text are converted into word sequences using a tokenization scheme that mainly relies on lowercasing the text, removing punctuations, and splitting the on whitespaces. These word sequences are then deterministically transformed into corresponding phoneme sequence representations, by relying on a lexicon of word pronunciations, in conjunction with a grapheme-to-phoneme model to approximate pronunciations for unknown words. Following prior work~\cite{serai2019improving,stiff2019improving}, we use the pronunciation lexicon provided as part of the Switchboard corpus~\cite{godfrey1992switchboard}, and use Phonetisaurus to train our grapheme-to-phoneme model on data from the same pronunciation lexicon. Special tokens such as noise, laughter, silence, end of utterance, were removed due to their absence in text data not of a spoken nature. A small number of examples (~2.2\%) in the Fisher data that contained zero words or phonemes in the ``true text'' as a result of this preprocessing were taken out prior to experimentation.

\section{Experiments and Intrinsic Evaluation}

\subsection{Training Details}
For the word level or single encoder model, we train the our network akin to a translation model using the Fairseq toolkit~\cite{ott2019fairseq}. For each pair of true and speech-recognized word sequences, the encoder is fed the true word sequence, and for each $i \in 1\ldots m$, we feed the decoder the first $i-1$ words from the speech-recognized sequence and give as a target the $i$th word of the speech-recognized word sequence, with a cross-entropy loss. We train with a Nesterov accelerated gradient descent~\cite{sutskever2013importance} optimizer for 60 epochs with a learning rate of 0.1 and a momentum of 0.99, with an optional 15 additional epochs in the finetune setting. To prevent overfitting, we employ a dropout of 0.2 on the input of every CNN layer, and on the output of the last CNN layer.

For the dual encoder model, we train our network similar to the word level model, except for two things. Firstly, Encoder A is fed the phoneme sequences corresponding to the true word sequence that is fed to Encoder B. Secondly, an Encoder Dropout of 0.5 is employed in addition to conventional dropout as used in the word level model i.e., on the input of every CNN layer, and on the output of the last CNN layer.


\subsection{Evaluation Metrics}
Following prior work~\cite{jyothi2009comparison,serai2019improving,serai2020end}, we use two metrics to evaluate the effectiveness of our models in hallucinating ASR errors, in addition to measuring the impact of our hallucinated ASR errors on the question classification model.

The first metric measures the percentage of real test set Error Chunks recalled in a set of ``K best'' simulated speech recognized utterances for each gold word sequence. The error chunks are again determined by aligning the gold word sequence with the errorful word sequence and removing the longest common subsequence. For example, if the gold sequence is ``\underline{do} you take any other \underline{medications} except for the tylenol for pain'' and the errorful sequence is ``you take any other \underline{medicine cations} except for the tylenol for pain,'' the error chunks would be the pairs \(\{medications:medicine\:cations\}\) and \(\{do:\:\:\}\). Our detection of error chunks is strict --- for an error chunk to qualify as predicted, the words adjacent to the predicted error chunk should be error-free.

The second metric measures the percentage of times the complete test set utterance is recalled in a set of ``K best'' simulated utterances for each gold text sequence (including error-free test sequences). We aimed to produce 100 unique simulated speech recognized utterances for each gold word sequence, so for both of these metrics, we evaluate the performance at K=100.
These are both ``hard'' metrics since the possibilities of various kinds of errors is quite endless, and the metrics only give credit when the utterance/error chunk exactly matches what was produced.

\begin{table*}[htb]
\centering
\caption{Evaluation on test sets of unseen Fisher corpus recognition data from the same recognizer. The {\em ConfMat} system predicts errors by sampling from a phone confusion matrix.  {\em Seq2Seq} Models directly predict the word sequences, and greatly outperform prior published results from {\em ConfMat} systems.}
\begin{tabular}{|c|c|c|c|c|}
	\hline
	\multirow{2}{*}{Model} & \multicolumn{2}{c|}{Error Chunks Predicted} & \multicolumn{2}{c|}{Complete Utterances Predicted}\\
	\cline{2-5}
	&Smaller Test Set&Full Test Set&Smaller Test Set&Full Test Set\\
	\hline\hline
    ConfMat w/ Direct decoding~\cite{serai2019improving} & 14.9\% & - & 39.2\% & - \\
	\hline
    ConfMat w/ Sampled decoding~\cite{serai2019improving} & 25.6\% & - & 38.8\% & - \\
    \hline
    Word level Seq2Seq w/ Beam Search decoding & 45.0\% & 43.4\% & \textbf{57.8}\% & 57.5\%\\
    \hline
    Word level Seq2Seq w/ Sampled decoding & 47.0\% & 46.2\% & 56.8\% & 57.5\%\\
    \hline
    Dual encoder Seq2Seq w/ Beam Search decoding & 47.6\% & 44.9\% & \textbf{57.8}\% & \textbf{58.2}\%\\
    \hline
    Dual encoder Seq2Seq w/ Sampled decoding & \textbf{48.8}\% & \textbf{47.1}\% & 56.2\% & 57.9\%\\
    \hline
\end{tabular}
\label{table-fisher}
\end{table*}

\subsection{In-Domain Evaluation}
\label{ssec:in_domain_evaluation}

In the in-domain evaluation setting, we measure our models' ability to predict errors on audio from the same corpus, and transcribed using the same speech recognizer, as used to generate their training data.

Table \ref{table-fisher} shows the results on the held out test sets from the Fisher corpus (\ref{sssec:fisher_base}) for our word level and dual encoder end to end models, comparing with our prior reported results using a confusion matrix based model on the smaller test set. Both the end to end models greatly improve over the previous best reported results with sampled decoding on the confusion matrix, in terms of real error chunks recalled, as well as complete speech-recognized utterances recalled. The dual encoder model outperforms the word-level end to end model on both metrics on the full test set, corroborating previous observations about the usefulness of phonetic information in improving generalization to words with limited or no examples in the training set.

The sampled decoding mechanism does the best on the error chunk prediction metric, which agrees with previous observations about the peaky nature of errors made by modern neural network based speech recognizers. However, it also brings a slight penalty on the complete utterance prediction metric, compared with the beam search decoding, perhaps because we sample the final output words independently for each time step, whereas beam search scores the word sequences by cumulative weight.

\begin{table}[htb]
\centering
\caption{Evaluation on out-of-domain read-speech Virtual Patient data from a cloud-based ASR service from 2018.  The same systems are used as in Table~\ref{table-fisher}. Our dual encoder models provide the highest fidelity to real ASR outputs.}
\begin{tabular}{|p{0.48\linewidth}|p{0.14\linewidth}|p{0.2\linewidth}|} \hline
	Model & Error Chunks Predicted & Complete Utterances Predicted\\
	\hline \hline
    ConfMat w/ Direct decoding~\cite{serai2019improving} & 8.5\% & 66.9\%\\
	\hline
    ConfMat w/ Sampled decoding~\cite{serai2019improving} & 36.4\% & 72.4\%\\
    \hline
    Word level Seq2Seq w/ Beam Search decoding & 39.2\% & 74.1\%\\
    \hline
    Word level Seq2Seq w/ Sampled decoding & 42.3\% & 74.2\%\\
    \hline
    Dual encoder Seq2Seq w/ Beam Search decoding & 41.9\% & \textbf{74.5}\%\\
    \hline
    Dual encoder Seq2Seq w/ Sampled decoding & \textbf{43.4}\% & 73.8\%\\
    \hline
\end{tabular}
\label{table-vp1}
\end{table}

\begin{table}[htb]
\centering
\caption{Evaluation on scare resource conversational-speech Virtual Patient data from a cloud-based ASR service from 2020. Systems that are finetuned on recognizer-specific data after training on unrelated recognizer data (Fisher\_base + Fisher\_finetune + VP\_conv), perform better than systems trained only on either type of data.}
\begin{tabular}{|p{0.2\linewidth}|p{0.33\linewidth}|p{0.12\linewidth}|p{0.19\linewidth}|} \hline
	Data & Model & Error Chunks Predicted & Complete Utterances Predicted\\
	\hline \hline
	\multirow{4}{*}{Fisher\_base} & Word level Seq2Seq w/ Beam Search decoding & 40.7\% & 82.0\%\\
    \cline{2-4}
    & Word level Seq2Seq w/ Sampled decoding & 47.0\% & 81.7\%\\
    \cline{2-4}
    & Dual encoder Seq2Seq w/ Beam Search decoding & 42.7\% & 82.6\%\\
    \cline{2-4}
    & Dual encoder Seq2Seq w/ Sampled decoding & 49.1\% & 82.3\%\\
    \hline
    \hline
    \multirow{4}{*}{Fisher\_base} & Word level Seq2Seq w/ Beam Search decoding & 72.3\% & 90.8\%\\
    \cline{2-4}
     & Word level Seq2Seq w/ Sampled decoding & 72.2\% & 90.8\%\\
    \cline{2-4}
    +Fisher\_finetune & Dual encoder Seq2Seq w/ Beam Search decoding & \textbf{75.0}\% & \textbf{91.5}\%\\
    \cline{2-4}
    +VP\_conv & Dual encoder Seq2Seq w/ Sampled decoding & 74.2\% & 91.4\%\\
    \hline
    \hline
    \multirow{4}{*}{Fisher\_finetune} & Word level Seq2Seq w/ Beam Search decoding & 65.9\% & 89.0\%\\
    \cline{2-4}
    & Word level Seq2Seq w/ Sampled decoding & 65.9\% & 88.9\%\\
    \cline{2-4}
    +VP\_conv& Dual encoder Seq2Seq w/ Beam Search decoding & 70.1\% & 89.0\%\\
    \cline{2-4}
    & Dual encoder Seq2Seq w/ Sampled decoding & 69.9\% & 89.3\%\\
    \hline
\end{tabular}
\label{table-vp2}
\end{table}

\subsection{Out-of-Domain and Scarce-Resource Evaluation}
\label{ssec:scarce_resource_evaluation}
In practice, we are hoping for error hallucination to help in the various scenarios where task-specific labeled speech data is limited or unavailable, and thus we also measure the quality of our models' hallucinated transcripts in out-of-domain and limited-resource settings.

Table \ref{table-vp1} shows the results on predicting recognition errors made by the cloud based ASR service from 2018 on the Virtual Patient read speech subset~(\ref{sssec:vp_read}), for comparing results to prior work. We use the same models from Table~\ref{table-fisher} so this is a completely out-of-domain evaluation where the recognizer as well as audio domains are unseen at train time. All our end to end models again improve on the best prior reported results on both error chunk and complete utterance prediction metrics, although the improvements are more modest in this case compared to the in-domain setting. In preliminary published work~\cite{serai2020end}, we reported how the output of the word level end to end model was different compared to the output of the phonetic confusion matrix model in this out-of-domain case, and the diversity of information gained from phonetics is again underscored again here by the gains seen due to use of the dual encoder model.

We also evaluate our models' ability to predict recognition errors seen on audio from the Virtual Patient conversational speech set~(VP\_conv), made by a recent 2020 version of a cloud-based ASR service. Table \ref{table-vp2} shows the results for predicting recognition errors on this set from our models trained in settings with zero as well as limited recognizer-specific ASR data available.

First, we evaluate base versions of our models i.e., the same as the ones evaluated in Tables~\ref{table-fisher} and ~\ref{table-vp1}, just trained on transcripts of the Fisher training set from an unrelated speech recognizer as compared to test time test-time~(Fisher\_base). Perhaps unsuprisingly, the results are comparable to what we see on the read speech data in Table~\ref{table-vp1}.

Further, we take the base versions of our models and train them further with the finetuning sets from the same speech recognizer as test-time, viz.: the Fisher finetuning set (Fisher\_finetune) and the train portion of VP Conversational Speech Set (VP\_conv) for up to 15 epochs. This results in an over 50\% relative increase in error chunk recall on this test set, and approximately 9\% absolute increase in complete utterance recall, showing a great benefit from the finetuning on recognizer-specific data including some domain-specific data.

As we see great benefit from finetuning,, we evaluate versions of our models that are train for 75 epochs only on the finetune sets i.e., data from the same recognizer as test time. While these models perform better than the base models trained only on unmatched recognizer data (Fisher\_base), they are not as good as the finetuned versions of the base models. 

Overall, in Table~\ref{table-vp2}, we find that our finetuned models that learn from both the larger but unmatched recognizer data (Fisher\_base), as well as the smaller but matched recognizer data (Fisher\_finetune and VP\_conv), perform better than those only trained on either of them. The dual-encoder architecture still does the best, showing the continued benefit of the phonetic representation. Surprisingly, unlike what we see with the base version of the models Tables~\ref{table-fisher}, \ref{table-vp1}, and the Fisher\_base rows of Table \ref{table-vp2}), the sampled decoding no longer helps improve error chunk recall on the finetuned models, in fact it hurts slightly. Our hypothesis for the cause behind this is that: with the Fisher\_finetune and VP\_conv sets, we are able to better model contextual errors resulting from the recent cloud based recognizer, and beam search's ability to consider the likelihood of sequences of words in the output outweighs the benefits of sampling that we see in other scenarios.

\begin{table*}[htb]
\centering
\caption{Extrinsic evaluation on conversational-speech Virtual Patient data. In the zero ASR data case, only typed or human annotated text were used for all training examples, whereas in the some ASR data case, real ASR transcripts from the ``Cloud-2020'' system were utilized for about half of the training examples (VP\_conv). Hallucination helps, particularly on higher WERs.}
\begin{tabular}{|c|c|c|c|c|c|}
	\hline
	\multirow{2}{*}{Test ASR system} & \multirow{2}{*}{WER} & \multicolumn{2}{c|}{\% Accuracy/F1 with zero ASR data} & \multicolumn{2}{c|}{\% Accuracy/F1 with some ASR data} \\
	\cline{3-6}
	 & & W/o hallucination & W/ hallucination & W/o hallucination & W/ hallucination\\
	\hline\hline
    Gold & 0\% & 79.7/59.8 & \textbf{79.8/60.2} & 80.4/60.3 & \textbf{80.6/60.4}  \\
	\hline
    Cloud-2020 & 8.9\% & 78.0/57.8 & \textbf{78.3/58.3} & 79.3/58.7 & \textbf{79.5/59.0}  \\
	\hline
    Cloud-2018 & 12.5\% & 76.1/56.6 & \textbf{77.4/58.0} & 77.0/57.4 & \textbf{77.8/57.8}  \\
    \hline
    Librispeech & 41.1\% & 65.6/45.8 & \textbf{67.9/48.6} & 66.7/47.3 & \textbf{67.5/48.5} \\
    \hline
\end{tabular}
\label{table-vp-downstream}
\end{table*}

\section{Extrinsic Evaluation}


In order to investigate the benefit of our hallucination approach to spoken language understanding, we perform an extrinsic evaluation on the Virtual Patient task. We use our models to simulate the effect of speech recognized input during the training of a question classification system, to see if they help alleviate degradations in performance caused by ASR errors in the input.

\subsection{Downstream Model}
\label{ssec:ques_class}
We use a self-attention RNN~\cite{lin2017structured} based question-classification model adapted for the Virtual patient task~\cite{stiff2020self}. This model uses a single layer BiGRU as the RNN. For the attention mechanism, we use 8 attention heads. Each attention head produces a representation of the input attending to different parts of the text. The vector representations from the 8 heads are then concatenated and fed into a fully connected classification layer with softmax activations to predict a distribution over 376 classes. Unlike the originally proposed model~\cite{lin2017structured}, we do not impose an orthogonality constraint on the attention scores from different attention heads. We found that doing so hurt the classification performance.

\subsection{Training and the use of Hallucination}
We train our model to minimize the cross-entropy loss using the Adam optimizer~\cite{kingma2014adam} with a learning rate of 0.001, a dropout rate of 0.5, and early stopping with a patience of 15 epochs. In the baseline case i.e., without error hallucination, the training uses the gold or typed versions of the text as input along with corresponding class labels. In the settings with some real ASR training data, the speech recognized versions of the input from the Virtual Patient Conversational Speech set are added to the training set.

In the error hallucination case, we use a sampling strategy~\cite{stiff2019improving} wherein, at train time, the input text for the question classifier is randomly transmuted with a pseudo-speech-recognized utterance sampled from the output of our finetuned ASR error prediction model (best one from Table~\ref{table-vp2}), except that in the zero domain-specific ASR data case the VP\_conv portion is excluded during finetuning. The sampling rate is treated as a hyperparameter and chosen from the set $\{5.0\%,10.0\%,25.0\%,50.0\%,75.0\%,100.0\%\}$ by tuning on the development set. A sample rate of $x\%$ means that the a training instance is replaced by a corresponding errorful alternative with a $x\%$ probability.

\subsection{Results}
Table~\ref{table-vp-downstream} shows question classification performance with and without ASR hallucination, to measure changes in Accuracy and Macro-F1 scores averaged across 5 random seeds. We observe that our proposed approach for hallucination helps improve downstream performance in multiple WER settings whether real ASR training data for the NLU task is available or not. We also observe that with increase in WER, the benefit from improvements from performing ASR hallucination can even be higher than using real ASR data. In the 12.5\% WER setting, adding hallucinated ASR in addition to some real ASR data, improved accuracy from 76.1\% to 77.8\% i.e., about twice as much as the improvement from real ASR data alone. In the 41.1\% WER setting, even with zero ASR data, our hallucination approach allowed an absolute 2.3\% improvement in downstream task accuracy, whereas real ASR data alone gave an improvement of 1.1\%. Notably, this shows that our hallucination approach can improve the NLU system performance even more than by adding some real ASR data. We reason that this happens because the use of real ASR data provided at most one alternative transcript containing ASR errors per training example, whereas our hallucination approach allows the model to see multiple plausible errorful transcripts per training example, potentially even a different one every epoch.

It is worth noting that error hallucination also improves performance slightly on gold transcripts which suggests that it acts like a soft data augmentation as proposed in~\cite{wei2019eda}. However, this improvement is not as high as that in noisy scenarios especially in higher WER settings.


\section{Conclusion and Future work}
We show that our sequence to sequence models greatly improve the error prediction performance over confusion matrix prediction approaches, which we attribute to their ability to model speech recognition behavior in a context dependent manner. We also observe that a combined use of phonetic and word level representations on input text through a dual encoder approach further improves the fidelity of its hallucination to actual behavior of the ASR system being characterized. With regards to sampling, which is a strategy that has helped improve error chunk recall in prior work, we found sampling to help when the characterized ASR system is out-of-domain or just simpler and trained on a single corpus. However, we think that our naive incorporation of it may be inhibiting the contextual model of the decoder network by taking away ability to search through full sequences, opening up the potential for future work, such as a variational sampling approach.

We also find that our ASR hallucination approach helps train a language understanding model to be robust to real ASR errors at test-time, and that the diversity of hallucinated ASR errors allow for an even greater benefit than training with some real ASR data in higher WER scenarios.


\section{Acknowledgements}
\label{sec:ack}
This material is based upon work supported by the National Science Foundation under Grant No. 1618336. We gratefully acknowledge the support of NVIDIA Corporation with the donation of the Quadro P6000 GPU used for this research. Additional computing resources provided by the Ohio Supercomputer Center~\cite{OhioSupercomputerCenter1987}. We thank Adam Stiff, Doug Danforth, and others involved in the Virtual Patient project at The Ohio State University for sharing various data from the project for our experiments. We thank Peter Plantinga and Speech Brain team for providing the trained Librispeech ASR model used.


%




\ifCLASSOPTIONcaptionsoff
  \newpage
\fi



%
\bibliographystyle{IEEEtran}
\bibliography{refs}



%








\end{document}